%% file: main.tex
\documentclass[sigconf,nonacm]{acmart}

\usepackage{multirow}

\usepackage{tikz}
\usetikzlibrary{positioning}
\usetikzlibrary{fit}
\usetikzlibrary{graphs}
\usepackage{float}
\usepackage{graphicx}
\usetikzlibrary{arrows.meta}
\usepackage{pgfplots}
\pgfplotsset{compat=newest}
\usetikzlibrary{decorations.pathmorphing, calc, backgrounds, positioning}
\usetikzlibrary{shapes.geometric}
\usepackage{xcolor}
\usetikzlibrary{shapes.multipart}
\usepackage{enumitem}
\usepackage{cleveref}
\usepackage{capt-of,etoolbox}
\usepackage{subcaption}
\usepackage{caption}

\begin{document}


\title{Detecting Multimodal Situations with Insufficient Context and Abstaining from Baseless Predictions}
\author{Junzhang Liu}
\authornote{Equal Contribution. The orders of the authors are based on rolling dice.}
\affiliation{%
  \institution{Columbia University}
  \city{New York}
  \country{USA}}
\email{jl6262@columbia.edu}

\author{Zhecan Wang}
\authornotemark[1]
\affiliation{%
  \institution{Columbia University}
  \city{New York}
  \country{USA}}
\email{zw2627@columbia.edu}

\author{Hammad Ayyubi}
\affiliation{%
  \institution{Columbia University}
  \city{New York}
  \country{USA}}
\email{ha2578@columbia.edu}

\author{Haoxuan You}
\affiliation{%
  \institution{Columbia University}
  \city{New York}
  \country{USA}}
\email{hy2612@columbia.edu}

\author{Chris Thomas}
\affiliation{%
  \institution{Virginia Tech}
  \city{Virginia}
  \country{USA}}
\email{chris@cs.vt.edu}

\author{Rui Sun}
\affiliation{%
  \institution{Columbia University}
  \city{New York}
  \country{USA}}
\email{rs4110@columbia.edu}

\author{Shih-Fu Chang}
\affiliation{%
  \institution{Columbia University}
  \city{New York}
  \country{USA}}
\email{sc250@columbia.edu}

\author{Kai-Wei Chang}
\affiliation{%
  \institution{University of California, Los Angeles}
  \city{Los Angeles}
  \country{USA}}
\email{kwchang@cs.ucla.edu}

\begin{abstract}

Despite the widespread adoption of Vision-Language Understanding (VLU) benchmarks such as VQA v2, OKVQA, A-OKVQA, GQA, VCR, SWAG, and VisualCOMET, our analysis reveals a pervasive issue affecting their integrity: these benchmarks contain samples where answers rely on assumptions unsupported by the provided context. Training models on such data foster biased learning and hallucinations as models tend to make similar unwarranted assumptions. To address this issue, we collect contextual data for each sample whenever available and train a context selection module to facilitate evidence-based model predictions. Strong improvements across multiple benchmarks demonstrate the effectiveness of our approach.
Further, we develop a general-purpose Context-AwaRe Abstention (CARA) detector to identify samples lacking sufficient context and enhance model accuracy by abstaining from responding if the required context is absent. CARA exhibits generalization to new benchmarks it wasn't trained on, underscoring its utility for future VLU benchmarks in detecting or cleaning samples with inadequate context. Finally, we curate a Context Ambiguity and Sufficiency Evaluation (CASE) set to benchmark the performance of insufficient context detectors. Overall, our work represents a significant advancement in ensuring that vision-language models generate trustworthy and evidence-based outputs in complex real-world scenarios.

\end{abstract}

\begin{teaserfigure}
    \captionsetup{type=figure}
    \begin{center}
    \captionsetup{skip=5pt}
    \includegraphics[width=1.0\linewidth]{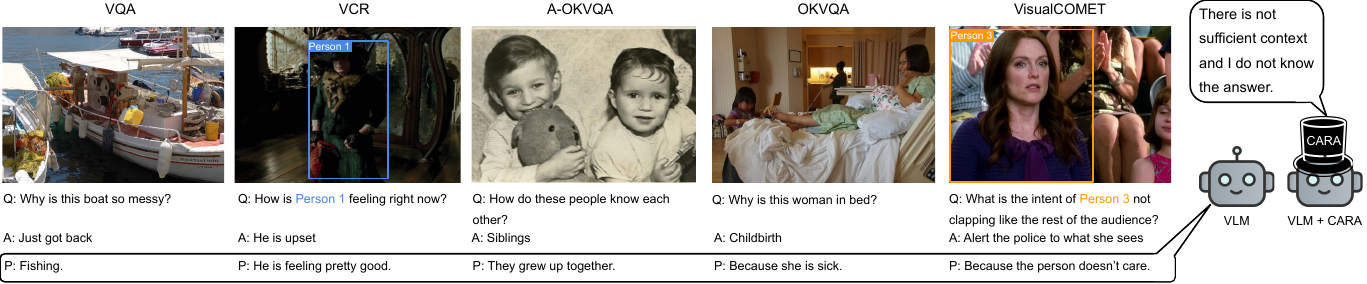}
    \captionof{figure}{
        Examples of samples with insufficient context to answer the question across several representative Vision Language Understanding (VLU) benchmarks. ``Q'' represents the question, ``A'' stands for the answer, and ``P'' denotes a typical Vision Language Model (VLM; here BLIP-2) prediction. We find that samples with insufficient context are common across several VLU benchmarks, causing VLMs to hallucinate predictions. Using (wearing) our proposed CARA (hat), VLMs are able to abstain from responding instead of making baseless predictions in such cases.
    }
    \label{fig:vlu_examples}
    \end{center}    
\end{teaserfigure}

\maketitle

\ccsdesc[500]{Computing methodologies~Natural language generation}
\ccsdesc[500]{Computing methodologies~Image representations}

\ccsdesc[500]{Computing methodologies~Natural language generation}
\ccsdesc[500]{Computing methodologies~Scene understanding}
\keywords{Context, Multimodal, Vision, Language}


\section{Introduction}
A number of Vision Language Understanding (VLU) benchmarks have been proposed to evaluate the capability of models to interpret complex multimodal scenarios and events \cite{okvqa, aokvqa, VCR, vqa, zellers2018swag, park2020visualcomet}. 
However, these benchmarks often include samples with insufficient event-specific context to answer the given questions.
For instance, in the first example of \Cref{fig:vlu_examples}, it is impossible to answer why the boat is messy without knowing what had happened before. 
Similarly, in the second example of \Cref{fig:vlu_examples}, knowledge of [Person 1]'s prior interaction is required to determine how the person feels. Answering the questions for these examples requires more contextual information about the events depicted in the images than is available from the image alone.


Our analysis reveals that this issue of insufficient event-specific context is pervasive in many VLU datasets. \Cref{fig:vlu_examples} illustrates examples from some representative benchmarks -- VQA v2 \cite{vqa}, Visual Commonsense Reasoning (VCR) \cite{zellers2019recognition}, OKVQA \cite{okvqa}, A-OKVQA \cite{aokvqa}, and VisualCOMET \cite{park2020visualcomet}.
The lack of sufficient and specific context in the provided samples forces models trained on such data to guess possible answers, leading to models that confidently predict answers without evidential support. 
Models that tend to hallucinate assumptions in this way undermine their trustworthiness and limit their real-world applicability in settings where accuracy is critical e.g., assistive technologies for the visually impaired \cite{web-automation, gurari2018vizwiz, zorec2020visual}, autonomous vehicles and robotics \cite{8967670, 10.1145/3337791}, healthcare applications \cite{lin2023medical} or security and surveillance \cite{toor2019biometric}. 


Our findings of the ubiquity of this problem lead us to two critical questions: 
1) If the context can be retrieved, \textit{e.g.}, we can obtain the corresponding video as context when the sample has an image from that video, how to identify the most necessary context and effectively incorporate it into models? 2) If there is no available context, \textit{e.g.}, the sample has an in-the-wild image, can we develop a generalizable method to identify samples with insufficient context and abstain from making baseless predictions?

Regarding the first question, 
numerous methods \cite{Lee2021VisionLanguageKnowledgeCF, song2020kvlbert} have been proposed to enhance image-text understanding with external knowledge.
Yet, these approaches fail to address the absence of \textit{event-specific} context, which is not available in external sources.
The challenges presented in Figure \ref{fig:vlu_examples}, for example, cannot be overcome simply through the application of general knowledge, as they require insights directly related to the depicted events.

As for the second question, no prior work has focused on abstaining from speculative responses by identifying insufficient event-specific context across existing VLU benchmarks. Existing works refrain from answering either due to low model confidence \cite{whitehead2022reliable} or due to out-of-distribution samples  \cite{Dancette_2023_CVPR}.
Consequently, they would still make unfounded predictions for samples with high model confidence or in-domain samples but insufficient context.

We address these limitations by 1) Collecting contextual data where available (VCR, SWAG, and VisualCOMET) and building a novel model-agnostic plug-and-play context selection module to incorporate context into model prediction (see \Cref{fig:example_of_3_dataset}); 2) Reusing the aforementioned module to collect pseudo-labels to train a Context-AwaRe Abstention (CARA) module, capable of identifying samples with insufficient context. Both our context selection module and CARA are model- and task-agnostic.


Our experiments demonstrate that our context selection module consistently improves performance across models and tasks. In the process, we also investigate several important questions: 1) In which modality (visual, textual, or both) does context benefit the most? 2) How long is context useful before it becomes noise? 
Further, we show that CARA boosts state-of-the-art model performance in full-shot and zero-shot settings by reducing inaccurate predictions through abstention from baseless or hallucinated responses. CARA, trained on one benchmark, can effortlessly generalizes to other benchmarks as well. This provides evidence that CARA could be useful for future benchmarks as well without any re-training. It could even be used to clean future benchmarks of samples with insufficient context.
Moreover, as CARA prevents models from making predictions that are not grounded in contextual evidence, we believe it will significantly improve model trustworthiness.

Lastly, to evaluate CARA's quality and benchmark its performance, we also curate an evaluation set manually annotated with the labels -- sufficient or insufficient context. This data is valuable for the future development of insufficient context detectors.

In summary, our work makes several key contributions:

\vspace{-0.7mm}

\begin{itemize}
    \item \textbf{Highlighting a Systemic Issue}: We identify a pervasive problem in common VLU benchmarks, i.e., the presence of samples with insufficient context. This issue has been largely overlooked in prior studies, despite its impact on the performance and reliability of VLU models. We conduct an extensive analysis across benchmarks to reveal the extent of this problem.

    \item \textbf{Incoporating Context Effectively}: We address the issue of insufficient event-specific context in VCR, SWAG, and VisualCOMET benchmarks by introducing a novel context selection method. This enhances model performance by accurately identifying and integrating relevant context into task resolution.

    \item \textbf{Multimodal Abstention Detector}: We develop CARA, a method for abstaining on samples lacking necessary context, and demonstrate its generalization across new benchmarks.

    \item \textbf{Data Contribution}: We collect contextual data for VCR, SWAG, and VisualCOMET, which is valuable for further exploration of context-aware model prediction. Moreover, we create a Context Ambiguity and Sufficiency Evaluation (CASE) set for insufficient context detection.
\end{itemize}

\input{graphs/examples}

\vspace{-1mm}

\section{Related Work}
\label{sec:related_work}
\subsubsection{Unanswerable Visual Questions}
The challenge of determining the answerability of visual questions has been explored primarily from two main directions: 
1) relevance of the question or 
2) quality of the image. 
The former direction focuses on creating datasets and methods that test models' ability to flag irrelevant questions \cite{kafle2017analysis, C2VQA, li2020neural, mahendru-etal-2017-promise, ray2016question} or questions inquiring about objects absent in the image \cite{lovenia2023negative, li2023evaluating, wu2023see}. On the other hand, the latter direction requires models to flag unanswerable samples due to low image quality \cite{gurari2018vizwiz, bhattacharya2019does}. Both directions overlook the nuanced complexity of unanswerability in the case of insufficient context for high-quality images paired with relevant questions. It is this gap that our work aims to bridge.  

\subsubsection{VLU with External Resources}When information in the image is insufficient to answer the question \cite{shahMYP19, okvqa, aokvqa}, several methods have been proposed to augment the provided information with external knowledge from Wikipedia \cite{lin2022retrieval}, the internet \cite{hu2023reveal}, and knowledge graphs \cite{song2020kvlbert, Lee2021VisionLanguageKnowledgeCF}. 
Our approach is similar in retrieving extra information to complement the provided visual information. 
However, we retrieve contextual information directly related to events and entities depicted in the image, while prior approaches search for general factual \cite{wikidata,dbpedia} or commonsense knowledge \cite{speer2018conceptnet, sap2019atomic}. 
The contextual information we seek, for example, the reason for [Person 1]'s injury in \Cref{fig:example_of_3_dataset}, is unavailable in those external sources.
Limited works have explored specific sample-related contextual information. \citet{Naik_2023_ICCV} utilize image source metadata while \citet{biten2019good,tran2020transform} leverage paired news article. Both cases bypass context retrieval by exploiting image metadata as is, unlike our work. Furthermore, they do not focus on integrating temporal or event-specific context, which is crucial for reasoning in semantically complex VLU tasks.


\subsubsection{Abstention in Multimodal Systems} 
Abstaining from responding instead of making incorrect predictions was originally explored in the unimodal language domain to address out-of-distribution or adversarial inputs \cite{chow1957optimum, Stefano827457, El-Yaniv2010, jiang2018pythia, kamath-etal-2020-selective, varshney-etal-2022-investigating}.
In the multimodal domain, recent works have been proposed that abstain similarly in the case of out-of-distribution samples \cite{Dancette_2023_CVPR} or low model confidence \cite{whitehead2022reliable}. In contrast, our proposed approach avoids making predictions when sufficient context to answer the question is unavailable. Unlike prior works, our abstention mechanism works from a data-centric view and applies to new benchmarks without any re-training.


\vspace{-1mm}

\section{Problem Space}
\label{sec:problem_space}

We investigate several benchmarks to study the problem of insufficient context in VLU domain: VQA v2, OKVQA, A-OKVQA, GQA, VCR, VisualCOMET, and Visual SWAG.
These datasets cover a range of VLU tasks, such as visual question answering, image-based text generation, and image-text matching. 
Notably, SWAG is a text-only entailed event inference dataset. To facilitate the study on multimodal event entailment inference, we replace the text premise in SWAG with the corresponding image frame retrieved from the samples' source video.
We call this multimodal dataset, Visual SWAG, where given an image premise, the required task is to infer the entailed event in textual form.

For the datasets with contextual data available -- VCR, VisualCOMET, and Visual SWAG -- we first collect that contextual data, as in \Cref{sec:dataset}. Then, we utilize it to facilitate evidence-based VLM prediction via a context selection module, as in \Cref{sec:context_selection}.
Further, we leverage a combination of vanilla VLM and VLM trained with context to pseudo-label samples with insufficient context. The pseudo labels are then used to train an insufficient context detector, CARA. We demonstrate that CARA generalizes to VQA v2, OKVQA, A-OKVQA, and GQA without having ever been trained on them.

The input information for the above reasoning benchmarks can be generally denoted as $x=(x_T,x_I)$ where $x_T$ is the textual input and $x_I$ is the image input. In our first study, we explore whether adding another input Context, $x=(x_T,x_I, C)$, can help and explore how to obtain the most beneficial context. For our second study, we 
develop functions to detect samples with insufficient context in the input, $x=(x_T,x_I)$, and abstain from making baseless predictions.


\vspace{-1mm}

\section{Contextual Data Collection}
\label{sec:dataset}

We begin by collecting contextual data for the three VLU benchmarks described above. These benchmarks evaluate models' understanding of events using images sourced from existing video datasets. To ensure comprehensive context is provided for each sample across the tasks, we collected multimodal contextual data, including preceding and subsequent visual frames along with paired text scripts.
We first discuss how we retrieved the source video data and then how context was retrieved and filtered. Finally, we present statistics about our assembled dataset.

\vspace{-2mm}
\subsection{Data Fetching}
\label{sec:data_fetching}

The images from VCR, VisualCOMET, and Visual SWAG are derived from video sources like LSMDC \cite{rohrbach2015dataset}, ActivityNet\cite{caba2015activitynet}, or YouTube. 
These video datasets consist of sequences of video clips, where each clip is paired with a sentence describing the event in the clip.
Since annotations in VCR, VisualCOMET, and Visual SWAG include specific frame IDs and clip IDs for most samples, we can locate and retrieve the source clip of the corresponding sample.
We removed all samples for which we could not find the corresponding source clip or paired video scripts.


\vspace{-2mm}
\subsection{Context Retrieval and Filtering}
\label{sec:context_retrieval}
\subsubsection{Context Retrieval} 
We retrieve the clips before and after the corresponding source clip as visual context.
These video clips are also paired with video scripts. We retrieve these scripts as text context for data points.
However, using video frames as visual context can be highly redundant due to their repetitive nature (i.e. adjacent frames are generally very similar), thus we find the most descriptive frame from each of these clips by finding the best match with the script using a pre-trained CLIP \cite{radford2021learning} model. 
More formally, we denote context from preceding clips with negative indices ($c_{-3}, c_{-2} ...)$, while context from succeeding clips has positive indices ($c_{1}, c_{2} ...)$, where each $c_i$ consists of both vision and language contexts.

\subsubsection{Context Filtering} 

Given that a substantial portion of our datasets comprises temporal questions, specifically those inquiring about states before and after, we take precautions to avoid inadvertently providing context that may disclose the answer to the model. We achieve this by identifying such cases using keywords and then filtering out contexts that could potentially lead to cheating. For instance, samples featuring questions about the past will be devoid of negatively indexed contexts.

\subsection{Data Statistics}
\label{sec:data_statistics}
\subsubsection{Training Data} Our training dataset split includes 41,008 image-text pairs from the train split of Visual SWAG, with an additional 94,404 distinct image-text pairs as multimodal context; 119,994 image-question pairs from the train split of VCR with 79,913 distinct image-text context pairs; and 190,457 image-text pairs (from 63,499 unique situations) from the train split of VisualCOMET, with 79,109 distinct image-text context pairs.

\subsubsection{Evaluation Data} Our validation set includes 10,645 image-text pairs from the train split of Visual SWAG with 57,880 distinct image-text pairs as multimodal context; 15,092 image-question pairs from the train split of VCR with 10,014 distinct image-text context pairs; and 23,930 image-text pairs (from 7,978 unique situations) from the train split of VisualCOMET, with 9,933 distinct image-text context pairs.


\input{graphs/method}

\section{Method}

Using our collected contextual data, we first develop a model-agnostic smart context selection module to add relevant context to samples to improve the model's understanding of the sample and, hence, its performance.
We then create a multimodal abstention model to identify samples lacking sufficient event-specific context and prevent models from making baseless predictions on such samples.


\subsection{Context Selection Module}
\label{sec:context_selection}

Consider a Vision-Language Model (VLM) \(M\), with image input $x_I$ and text input $x_T$. The most straightforward way to incorporate context, $C=[c_{i}]_{i \in [-n, n]}$, is to append it to the model input. That is, \(y_{pred} = M(x, C)\), where $x=(x_I, x_T)$.
This results in a brute-force context injection approach, which is both heavily computationally expensive and potentially noisy. 
Instead, 
we aim to build a method to intelligently select the most relevant context according to the given target image and text premises.

We thus propose a ``probabilistic context selection'' method (see Figure \ref{fig:method_fig}). 
This end-to-end, model-agnostic technique aims to streamline the selection of event-specific context. Our method features a context selection module
\(M_c\) designed to identify the most 
relevant context \(c^*\) for the given input $x$. 
As a result, the model's output is given by \(y_{pred}^* = M(x, c^*)\). 
The core idea behind this approach is that it can dynamically select the context that is most aligned with the input.
This allows it to integrate only the most relevant context into the downstream reasoning process while filtering out noisy context. We demonstrate that this significantly improves the model's ability to handle complex reasoning tasks requiring contextual information.

Specifically, for a given input \(x\) and a set of context \(c_i\), the selection module \(M_c\) computes a score vector \(S = [s_i]_{i \in [-n, n]} = M_c(x, c_i)\). Each score \(s_i\) within this vector denotes the relevance of \(c_i\) to \(x\).
The $c_i$ with the highest $s_i$ is chosen as the selected context for inference. During training, each $s_i$ is used to (softly) select the $c_i$ as the context in the VLM, $M$. 
This encourages the context selection module $M_c$ to assign a low weight to context $c_i$, which leads to a high loss in $M$ and vice versa. Thus, 
$M_c$ is trained to 
assign a high weight to the most relevant context.
The resulting loss function is:

\begin{equation}\small
\mathcal{L} = \sum_{i=-n}^n s_{i} \cdot l(M(x, c_i), y)
\label{eq:loss}
\end{equation}
where, \(l\) represents the cross-entropy loss.
 
This probabilistic sampling procedure, where $c_i$ is sampled using $s_i$, is differentiable end-to-end.
We illustrate how the context selection module interacts with the backbone VLM in Figure \ref{fig:method_fig}.

For a given input \(x\) and a specific context \(c_i\), we 
append the context with the input to create $x_i = [x \parallel c_i]$. More specifically, text context $c_{i_T}$ is appended to text input $x_T$ and image context $c_{i_I}$ with image input $x_I$, creating $x_{i_T} = [x_T \parallel c_{i_T}]$ and $x_{i_I} = [x_I \parallel c_{i_I}]$ respectively. $x_{i_T}$ and $x_{i_I}$ are then processed by $M$ as it would normally process $x_T$ and $x_I$.
 

\subsection{Multimodal Abstention Detector}
\label{sec:mm_abstention}
The above section assumes additional context is available to be recovered through retrieval. 
However, in many real-world scenarios, additional context may not be available. Thus, we propose a generalized multimodal abstention detector that aims to identify if a sample is unanswerable due to a lack of context thereby preventing baseless predictions.

Developing a mechanism to detect samples with insufficient event-specific context is an extremely challenging problem because the model must first hypothesize what the sufficient context would be to answer the question before determining if that context is lacking. 
In this work, we present a straightforward yet effective solution to address this issue. 
We leverage our previously trained model with context and compare its response with a vanilla model trained without context to pseudo-label if the sample contains or lacks sufficient context.
Our key insight is that if a sample already has sufficient context, the model's response should remain relatively consistent when additional context is added. Conversely, if the sample lacks sufficient context, the model's response should improve on adding additional context.
The pseudo labels are then used to train our insufficient context detector.
We illustrate this process in \Cref{fig:cimad} and detail it below.

\subsubsection{Confidence-Driven Pseudo-Labelling} 
\label{sec:pseudo-labelling}

We train two models: a Context-VLM (C-VLM), which incorporates context into its decision-making process (as detailed in \Cref{sec:context_selection}, and a vanilla VLM, which operates without context.
We compare the responses from both models to pseudo-label samples as follows:

\begin{itemize}

\item \textbf{Positive}: Instances correctly answered by the C-VLM model with high confidence above a designated threshold, $\gamma$, but incorrectly answered by the VLM with low confidence below a designated threshold $\mu$. 
The significant difference in accuracy and confidence suggests that these instances previously lacked sufficient context for unambiguous understanding.

\item\textbf{Negative}: Instances correctly recognized by both models with confidence above threshold $\gamma$, implying that context does not play a critical role in their identification.

\item\textbf{Excluded}: Instances not fitting into the above categories are excluded. 
The impact of context on these instances is uncertain, and including them could introduce noise into the model training process.
\end{itemize}

When pseudo-labeling the training set, we divide the dataset into two equal parts. Each part is labeled based on the inference results obtained from the two models trained on the 
other half. 
This strategy ensures a robust pseudo-labeling process that mitigates overfitting risks and data leakage from the validation set.

\subsubsection{Training. } After pseudo-labeling the data points, we train CARA, a 24-layer cross-modal attention module as our insufficient context detector via cross-entropy loss. 
The input to CARA is a sample image and the corresponding question or statement, and the output is a binary label denoting whether the sample lacks sufficient context or not.

\subsubsection{Inferencing. }
When inferencing using CARA, we predict if a data point lacks sufficient context based on whether CARA's prediction score exceeds a dataset-specific threshold $\theta$.

Note that during both training and inference, the detector operates \textit{without access to context}. Our goal is to develop a generalized detector that maintains high performance and generalizability across datasets, regardless of whether context is available. 


\input{graphs/cara}
\input{graphs/window_size_selection_num/window_size_selection_num}

\section{Experiments}
In this section, we present experimental results and analysis to illustrate the effectiveness of our context selection methodology and CARA. We first present implementation details, followed by context selection results and abstention detection results.

\subsection{Implementation Details}

\subsubsection{Base Models}
We demonstrate the efficacy of our approach on two different classes of VLMs: discriminative (VL-BERT\cite{su2020vlbert}) and generative (BLIP \cite{li2022blip}, BLIP2 \cite{li2023blip2}, MCAN\cite{yu2019mcan}, MiniGPT-4 \cite{zhu2023minigpt4}, OFA \cite{wang2022ofa}, PNP\cite{tiong2023plugandplay}, Prophet\cite{shao2023prompting}, and PromptCap\cite{hu2022promptcap} ).
Generalization 
across models shows that our approach is model-agnostic. 
We adhered strictly to the implementation described in the original papers and repositories of all models.

\subsubsection{Training}
Fine-tuning of VL-BERT, BLIP, and OFA is done with 2 NVIDIA-RTX 24 GB GPUs with batch size 32, BLIP2 and MiniGPT-4 are trained with 2 NVIDIA A100 GPUS with the same batch size. The initial learning rates for VL-BERT, BLIP, BLIP2, MiniGPT-4, and OFA are 7e-5, 1e-5, 2e-5, 3e-5, and 3e-5, respectively. VL-BERT is trained for 20 epochs, and BLIP is for 10 epochs, while BLIP2 and MiniGPT-4 are trained for 5 epochs. For OFA, we follow the original implementation
and train a total of 40K steps. Training of the models takes $\sim$48 hours. The abstention detector is trained for 10 epochs with a learning rate of 7e-5. 

\subsubsection{Context Selection}
BLIP, BLIP2, MiniGPT-4, and OFA lack native RoI functions as in VL-BERT. Thus, to process datasets requiring RoI handling such as VCR and VisualCOMET we adopt Merlot's approach \cite{zellers2021merlot} of drawing colored highlights around referenced entities in pixel space, as shown in Figure \ref{fig:example_of_3_dataset}.
In our experiments, we employ a Sentence-BERT \cite{reimers-2019-sentence-bert} as the text encoder for $M_C$, and a ViT \cite{dosovitskiy2021image} as the vision encoder. 
We fuse the global embeddings from those two encoders' output via concatenation and apply an MLP with sigmoid to map the fused feature into a score ranging from 0 to 1. 

\subsubsection{Abstention Detector} 
We use a 24-layer cross-modal attention model as the multimodal abstention detector, following \cite{su2020vlbert} to initialize and train it on datasets labeled with the pseudo-labeling method described in Section \ref{sec:mm_abstention}.
However, this results in an unbalanced training set with significantly more negative data points. 
To address this, we apply loss weighting during training. Our experiments show increasing the weight of positive data points in the loss by six 
results in the highest evaluation performance.

The CARA tailored for the Visual SWAG and VCR tasks are trained on their respective datasets. However, since VisualCOMET is a generative dataset without a binary correctness measure, we utilize the CARA trained on VCR for it instead. In practice, we utilize heuristics rules \footnote{Please refer to the Supplementary Materials for the detailed implementation steps.} integrating the confidence scores of the VLM and CARA's prediction to obtain the best model performance over downstream tasks. 

When conducting confidence-driven pseudo-labeling, the hyperparameters for filtering thresholds, $\gamma$, is 0.7 and $\mu$ is 0.5.

\subsection{Context Selection Results}
\label{sec:context_selection_results}

To determine the best way to integrate context into VLU tasks, we first perform extensive ablation experiments. We analyze various components of our context selection method, including modalities, window size, number of selected frames or scripts, and selection strategies. Finally, we apply our method to benchmark approaches.


\input{tables/context_modality}
\input{tables/context_model_result.tex}
\input{tables/case}
\input{tables/abstain}
\input{tables/abstain_generalization_v1}

\input{tables/abstain_generalization_v2}

\input{tables/risk}

\subsubsection{Data Modality Ablation} 
To ensure the best context utilization, we examine which modalities are most effective for both selecting relevant context and integrating it into VLMs. 
Table \ref{table:multimodal_context} shows results from experiments over VL-BERT with different input and output modalities for our context selection module.
The ``Selection modality'' column denotes the modality used to select context and ``Context Modality" refers to the modality of selected context \footnote{Please refer to the Supplementary Materials for details. \label{footref:appendix}}.
We find that using text as both the selection and context modality is the most effective approach. This trend holds across different context modalities (image, text, or both).



Regardless of the selection modality, we find adding visual context typically leads to a performance drop. Using text alone for context consistently yields the best results. While the visual context may offer rich information, our findings suggest it often introduces noise, which hurts performance. This highlights an opportunity for future research in integrating multimodal context in VLU tasks.  However, one critical finding is that our approach never decreases the performance of the VLM, even in the absence of text for selection or context.
To perform this ablation, we use a window size of 3, select one context unit, and use our probabilistic selection method.



\subsubsection{Window Size Ablation}
We next experiment with different context window sizes for VL-BERT on the Visual SWAG and VCR datasets. In this experiment, we limit the number of selected context units to 1, and the window size can range from 0 (no context) to 7. Figure \ref{fig:window_size} shows how VL-BERT's performance varies with different window sizes.
Our results indicate that models with nonzero window sizes outperform the baseline (window size of 0). However, performance plateaus and eventually decreases with excessively large window sizes. The peak accuracies on both VCR and Visual SWAG datasets suggest that their optimal window sizes are 3.


\subsubsection{Selection Number Ablation. }
Next, we analyze the impact of the amount of context on model performance.
Figure \ref{fig:selection_number} presents the results of training VL-BERT, BLIP2, and MiniGPT-4 on VCR and Visual SWAG datasets with different numbers of selected contexts over a window size of 3.
In this setup, \(M_c\) considers all possible combinations of concatenating \(r\) context from \(n\) available options arranged temporally rather than being limited to a single optimal context.
The models achieved their best performance across all three datasets with a selection number of 2. Notice the drop at the right end of each graph, where the selection number equals the window size. 
This extreme scenario inputs all the contexts inside the window without a context selection module and shows the importance of our selection module for improving context utilization.


\subsubsection{Selection Strategy Ablation. }

In Table \ref{table:selection_method}, we compare context selection strategies with VL-BERT to determine the most effective one. 
The bottom of the table presents results from heuristic methods based on context indices, while the top part explores dynamic selection strategies leveraging the embedding similarity. More specifically, we can rely on sentence similarity between the question and text context using Sentence-BERT \cite{reimers-2019-sentence-bert} after determining textual modality as the optimal selection modality. Both the embedding similarity method and heuristic methods are notably outperformed by our jointly trained model, the probabilistic context selection method. In this comparison, selection methods are limited to a window size of 3 and 2 selected contexts.


\subsubsection{Benchmark Comparison. } 


We apply our probabilistic context selection approach to various base models and report the results in Table \ref{table:context_model_result}. With our probabilistic selection method (+ Prob.), all five base models can generally improve their performance across three tasks. Furthermore, the base models can achieve SOTA scores on VisualCOMET with our selection method.
These results verify the benefits of incorporating contextual information into VLU tasks and the effectiveness of our method. 

\subsection{Abstention Detector Results}
In this section, we discuss the effect of our abstention detector by comparing the performance of VLMs with and without CARA.

\subsubsection{Evaluation of Detection Accuracy for Samples with Insufficient Context. }
\label{sec: cara-verify}
Building on the confidence-driven pseudo-labeling method outlined in Section \ref{sec:pseudo-labelling}, we assembled a small data pool of 500 positive and 500 negative image-question pairs from the VCR validation set and a similar one from Visual SWAG. These datasets were evaluated by Amazon Mechanical Turk workers to ascertain their ambiguity \footref{footref:appendix}. With this curated data, we created the Context Ambiguity and Sufficiency Evaluation (CASE) Set, spanning both benchmarks to evaluate the efficacy of abstention methods in detecting samples with insufficient context.

We compare CARA to two established methods \cite{whitehead2022reliable}: Selector-MaxProb, which abstains based on a predefined confidence threshold, and Selector-MLP, which predicts the likelihood of correct predictions using a Multilayer Perceptron module from an image and question. As demonstrated in Table \ref{table:case}, CARA exhibits high detection superior accuracy across these evaluation sets. Notably, when trained with the pseudo-labeled data from VCR, CARA also demonstrates strong performance on the CASE set of Visual SWAG, underscoring its generalizability. Moreover, the gap between CARA's performance and human judgment (oracle accuracy) underscores the ongoing challenges in detecting samples with insufficient context and highlights the value of the CASE set for future research.

\subsubsection{Performance Enhancement with CARA}
We compare the performance of baseline VLMs with and without CARA across three VLU tasks in Table \ref{table:abstention1}. When using CARA (+CARA), we calculate the accuracy of the baseline VLM only in instances where CARA indicates sufficient context for accurate prediction.
The results show that CARA results in a significant improvement in performance across all three tasks. 
Surprisingly, adding CARA can approach or even exceed the benchmarks set by context-aware models (referenced in Table \ref{table:context_model_result}), showing that CARA adds substantial value for multimodal abstention.

\subsubsection{Generalization Across VLU Benchmarks}
To test CARA's generalizability, we trained it on VCR and evaluated on VQA v2, GQA, OKVQA, and A-OKVQA in \Cref{table:abstain_gen1}.
Our findings demonstrate CARA's robust generalizability, emphasizing its significance for future VLU benchmarks for performance gains or insufficient samples cleaning.


If we further investigate the data points filtered out by CARA and examine them by humans\footnote{Please refer to the Supplementary Materials for the detailed implementation steps.}, as in table \ref{table:abstain_gen2}, we can observe that the majority of data points filtered out by CARA consist of ambiguous questions and most of them lack sufficient context for a determined answer. 
These results demonstrate CARA's effectiveness in enhancing model performance across benchmarks and underscore the key problem of instances lacking sufficient context within these benchmarks \footnote{The verification results of abstained samples by human review over VCR, VisualCOMET, and Visual SWAG can be found in the Supplementary Materials.}.

\subsubsection{Benchmarking Risk and Coverage}

Previous abstention strategies or selective prediction systems \cite{Stefano827457, El-Yaniv2010} were designed to optimize the balance between risk and coverage, where risk refers to the error rate for the predictions made, and coverage quantifies the total number of predictions issued. An optimal abstention strategy aims to minimize risk while maximizing coverage to the greatest extent possible. 
We assess the risk ($\mathcal{R}$) and coverage ($\mathcal{C}$) performance metrics of CARA compared to previous abstention strategies. Our goal is to minimize risk while maximizing coverage. We also evaluate the effective reliability ($\Phi_c$) of CARA, rewarding accurate predictions and penalizing incorrect responses.


Table \ref{table:risk_coverage} presents the evaluation results with varying risk tolerance levels (i.e.~how much risk a model accepts before abstention). 
We observe that in most cases CARA's $\mathcal{R}$'s system risk is roughly on par with existing methods while achieving significantly higher reliability and effective coverage for both BLIP2 \cite{li2023blip2} and PNP \cite{tiong2023plugandplay}.


\definecolor{cap_chosen}{rgb}{0.4752, 0.6273, 0.84}
\definecolor{no_cap_choice}{rgb}{0.7, 0.13, 0.13}
\definecolor{cap_choice}{rgb}{0.52, 0.73, 0.4}
\section{Qualitative Examples}
\input{graphs/qualitative_examples}
Figure \ref{fig:qualitative_examples} shows qualitative examples of effective context in VLU (top) and context-aware abstention (bottom).

\section{Conclusion}
In this paper, we discussed the issue of insufficient context grappling existing VLU benchmarks and proposed strategies to effectively integrate context, when available, or abstain from speculative prediction in case of samples with insufficient context. We also contributed datasets to enable further exploration of this problem.

\section{Limitation}
Please refer to the LIMITATION section in the Supplementary Materials for a detailed discussion.



\bibliographystyle{Reference-Format}
\bibliography{main}
\newpage
\appendix

\input{graphs/context_example}

\section{Importance of Context}
Context provides critical information to explain situations, avoid misinterpretations, and leverage fine-grained knowledge for prediction. It is particularly important in visual language understanding. For example, the ambiguities in \Cref{fig:context_importance} cannot be clarified without context. Lack of sufficient context can harm model learning and performance evaluation. However, ensuring adequate context exists in multimodal inputs with images and text is challenging and impractical for real-world scenarios, where additional context might not be available. Thus, the ability to abstain when needed context is missing is equally crucial.
\input{graphs/cara-verification}

\section{Additional Implementation Details}
\subsection{Heuristics with Context-Aware Abstention}
Since our method is data-centric and does not base its predictions on the output of the Vision Language Model (VLM), when deciding whether to abstain from an answer generated by a VLM, to account for the VLMs' variance, we combine the VLM's confidence with the prediction of the Context-AwaRe Abstention (CARA) detector according to a heuristic rule:
\begin{equation}
H = w(1-C)+(1-w)V
\label{eq:heuristic}
\end{equation}
\noindent where $V$ is the VLM's confidence, $C$ is CARA's confidence, and $0<w\leq1$ is the weighting of CARA's score.
A high $C$ indicates CARA predicts a need for context, so $1-C$ represents CARA's confidence in that the data point's has sufficient context. We use the heuristic score $H$ and a risk tolerance threshold to decide on abstaining or answering. This heuristic incorporates both CARA's and VLM's confidence scores via a weighted sum.


\section{Additional Ablation Details}
\subsection{Context Modality} As introduced in Section 6.1.3 of the main paper, for the context selection module, we encode image context and text context using ViT \cite{dosovitskiy2021image} and Sentence-BERT\cite{reimers-2019-sentence-bert}, respectively. The two embeddings are combined and passed through a Multilayer Perceptron to obtain the final score. The context is inputted to VLM by appending the image/text context to the input sequence. Thus, we can control the modality of context the VLM can observe by appending the corresponding contexts to the inputs. Similarly, the modality the context selection module uses to select context can also vary by adding/removing the vision or language encoder. For instance, when using text to select text-only context, we append only the text context to the input sequence for the VLM, and we only use the embeddings from Sentence-BERT for the context selection module.
\input{graphs/context_retrieval}
\section{Data Collection}
\subsection{Context retrieval}
The data points in  VCR, VisualCOMET, and Visual SWAG are sourced from either ActivityNet \cite{caba2015activitynet}, LSMDC \cite{rohrbach2016movie}, or YouTube.
Since only LSMDC data points have consistent and ordered context information available, we initially remove all non-LSMDC sourced data points in the Data Filtering stage, as depicted in \Cref{fig:context_retrieval},

In the Context Retrieval stage, we first sort the clips temporally. Then, we locate the source LSMDC clip for each QA data point. 
The script of the source clip serves as the text context of \(c_0\). The corresponding vision context is collected by finding the most relevant frame using a pre-trained CLIP \cite{radford2021learning} model, as mentioned in the main paper. 

The contexts at positive and negative indices are acquired with a similar procedure. For the context \(c_{\pm n}\), we traverse \(n\) clips forward or backward and apply the procedure mentioned above. Collecting all the \(c_{\pm n}\) will result in a context window size of $2n + 1$.

We set the maximum $n$ to be 20. This means each data point will include a range from \(c_{-20}\) to \(c_{20}\), totaling 41 context data sourced from LSMDC. We believe this adequately encompasses the necessary context for each question. Given that the average duration of LSMDC clips is 4.16 seconds, these 41 contexts collectively span approximately 2 minutes and 56 seconds of content.

Finally, in the Context Filtering stage, we remove the potentially cheating contexts for temporal questions by matching the keywords in the question. For example, \Cref{fig:context_retrieval} shows contexts with positive indices are removed for questions asking about ``After" to prevent the answer from leaking.

\input{tables/abstain_generalization_appendix}

\subsection{DATA QUALITY CONTROL FOR CASE}
Building on the confidence-driven pseudo-labeling method (Section 5.2.1), we assembled a small data pool of 500 positive and 500 negative image-question pairs from the VCR validation set and Visual SWAG. 
With this curated data, we created the Context Ambiguity and Sufficiency Evaluation (CASE) Set, spanning both benchmarks to evaluate the efficacy of abstention methods in detecting samples with insufficient context.
We evaluated these samples by Amazon Mechanical Turk workers to assess their ambiguity. We implemented the interface layout shown in \Cref{fig:cara-verify} and hired experienced annotators to manually verify the filtered samples. For each sample detected as positive (lacking sufficient context) by CARA, four experienced annotators re-verified it. The annotators were not informed of CARA's prediction and answered two curated questions independently. Based on the annotation results, we calculated the voting percentage to determine if each question was considered ambiguous and lacking sufficient context.
To ensure annotation consistency, we used Fleiss' Kappa ($\kappa$) \cite{Falotico2015FleissKS} to assess inter-annotator agreement. For determining if the question is ambiguous, $\kappa$ is 0.81, and for determining if the question lacks sufficient context, $\kappa$ is 0.84.

\section{Addtional Experiments and Results}

\subsection{Abstention Results Verification}
In Tables 4 and 5 of the main paper, we can observe that adding CARA on top of base VLMs can generally improve the performance across benchmarks. To further verify CARA's effectiveness and ensure that CARA focuses on removing problematic ambiguous samples (including samples with insufficient context) instead of challenging but answerable ones, we conduct manual human verification to examine the filtered-out data by CARA. Specifically, we let human annotators verify 100 randomly sampled instances for each dataset where CARA predicts positive (i.e., need context). In \Cref{table:abstain_gen_appendix}, we show human verification results on different datasets. We label ``ambiguous" for data points that have no obvious correct answer, as shown in the examples in \Cref{fig:abstained} of the supplementary materials. The ambiguity of these questions may vary. For example, the first question's reference to laptops is ambiguous since there is more than one brand in the image, and some cannot be determined due to poor image quality. Among these, a significant portion of ambiguity is caused by insufficient context, which happens when the question is ambiguous. Still, such ambiguity can be alleviated when additional information about the scene (i.e., context) is provided. Examples of this type are shown in Figures 1, 2, and 6 of the main paper, as well as highlighted in \Cref{fig:abstained}. We are surprised to find that CARA is able to identify other types of samples with ambiguities as well, such as those with ambiguous questions or poor image quality.

\input{graphs/appendix_qualitative_example}
\subsection{Qualitative Examples}
\subsubsection{Context Selection}
In \Cref{fig:q_examples_swag_vcr} and \Cref{fig:q_examples_visualcomet}, our contextual model demonstrates superior performance over the non-contextual model across numerous instances. Take, for instance, the third example from the Visual SWAG dataset. Without context, the correct choice, A, appears arbitrary, leading to the model incorrectly selecting choice D. However, our contextual model effectively identifies and leverages the relevant context—``someone gets up and goes over to the cool box''—to correctly associate it with the answer ``returns with four cans''. 

\subsubsection{Abstention}
\Cref{fig:abstained} shows the prediction of CARA, with the abstained samples labeled with ``Ambiguous" or ``Insufficient Context" by humans. We also provide BLIP2's response to these questions. Compared to the non-abstained questions (bottom two), the abstained ones have significantly diverse answer references, indicating disagreement among annotators. 

\section{Limitation}
Although CARA can be adapted to different problems and VLMs without needing to be retrained, the decision threshold and parameters for the heuristic rule in \Cref{eq:heuristic} may require additional tuning to achieve optimal performances.

The context selection method defined in Section 5 of the main paper works only for segmented contexts, which in our case consists of short sentences and videos. However, when applying it in other scenarios, for example, when context is in the form of paragraphs, context needs to be broken into pieces to adapt our method. In addition, the loss function mentioned in Section 5.1 of the main paper requires the model to recompute the input $m$ times given the context window size of $m$. This raises scalability issues for large context window sizes.

\end{document}

%% file: graphs/examples.tex
\begin{figure*}[h!]
    \centering
    \captionsetup{skip=5pt}
    \scalebox{0.7}{
    \includegraphics[width=\linewidth]{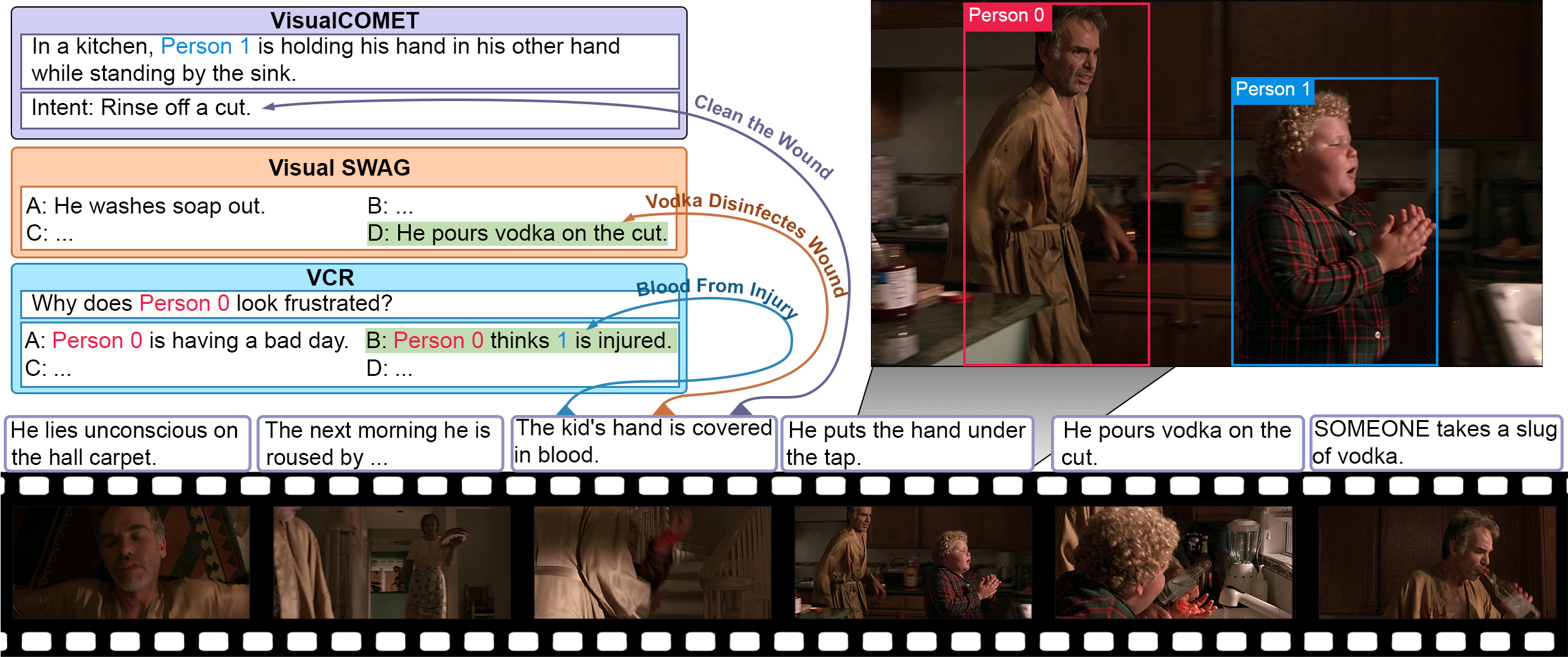}
    }
    \caption {
    Illustration of how we obtain contextual data for VCR, Visual SWAG, and VisualCOMET. The video from which the image sample is sourced is identified to obtain temporal context in the form of frames and captions around the image sample in question. The context provides the necessary evidence required to answer these highly semantic questions.}
    \label{fig:example_of_3_dataset}
    \Description[Illustration of our context dataset and how context can be helpful.]{Examples of data lacking context and how context helps answer them.}
\end{figure*}

%% file: graphs/method.tex
\begin{figure}[h!]
    \captionsetup{skip=5pt}
    \scalebox{0.8}{
    \includegraphics[width=\linewidth]
    {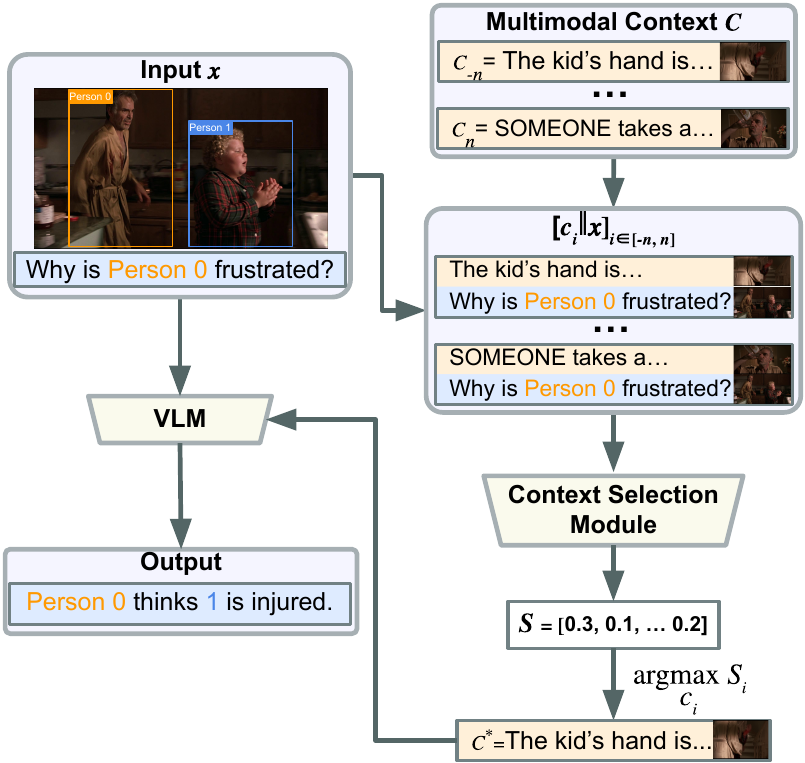}
    }
  \caption{A high-level demonstration of the probabilistic context selection method. For the VLM's input, in addition to the question and image, a context sentence selected by the Context Selection Module is appended to the original input.}
  \Description[Pipeline of the probabilistic selection method.]{The context selection method uses a selection module to generate a score for each context to select the most relevant context for the VLM to answer the question.}
  \label{fig:method_fig}
\end{figure}

%% file: graphs/cara.tex
\begin{figure}[t]
    \centering
    \captionsetup{skip=5pt}
    \scalebox{0.8}{
    \includegraphics[width=\linewidth]{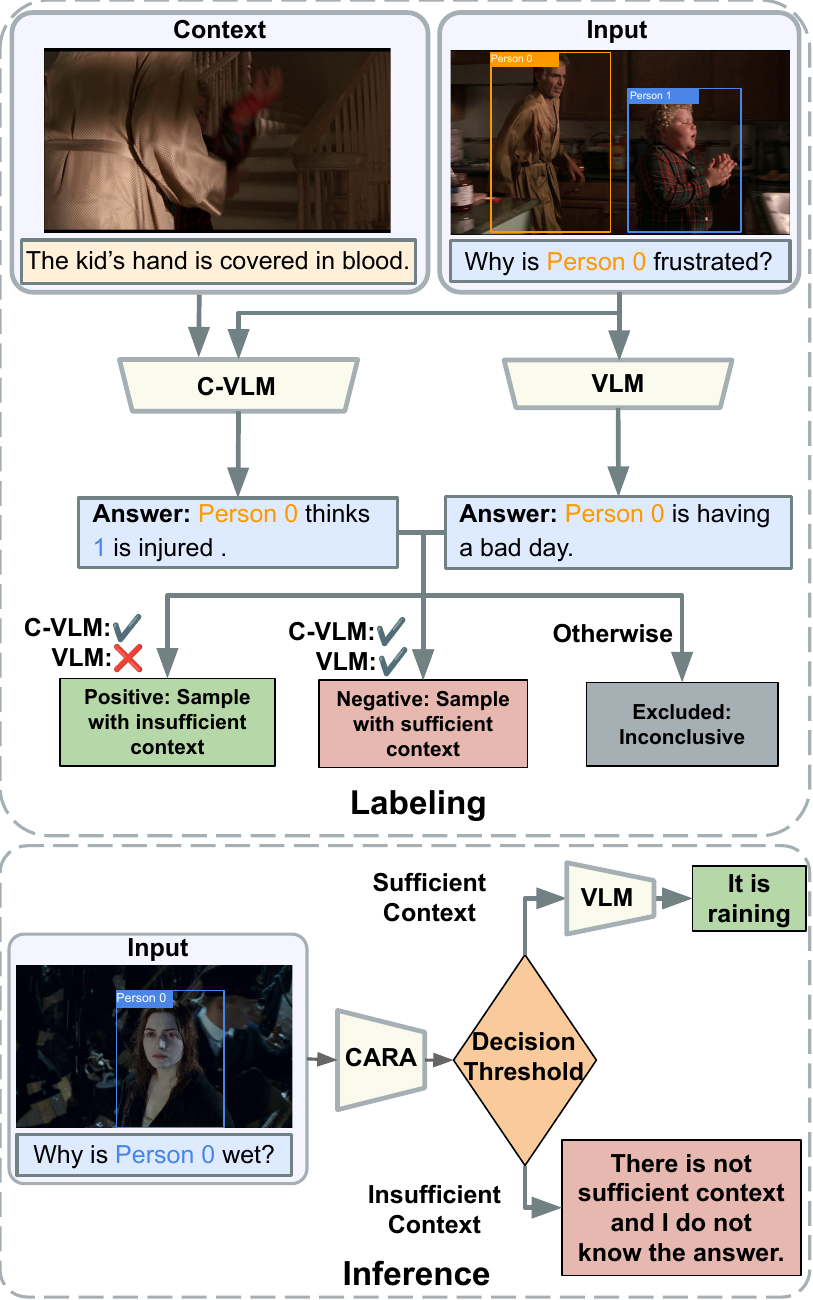}
    }
    \vspace{-0.5em}
  \caption{Top: We use models with/without context to pseudo-label whether instances need context.
  The labeled data is then used to train CARA. Bottom: CARA decides whether to abstain based on whether the input contains sufficient context.}
  \label{fig:cimad}
  \vspace{-1em}
\end{figure}

%% file: graphs/window_size_selection_num/window_size_selection_num.tex
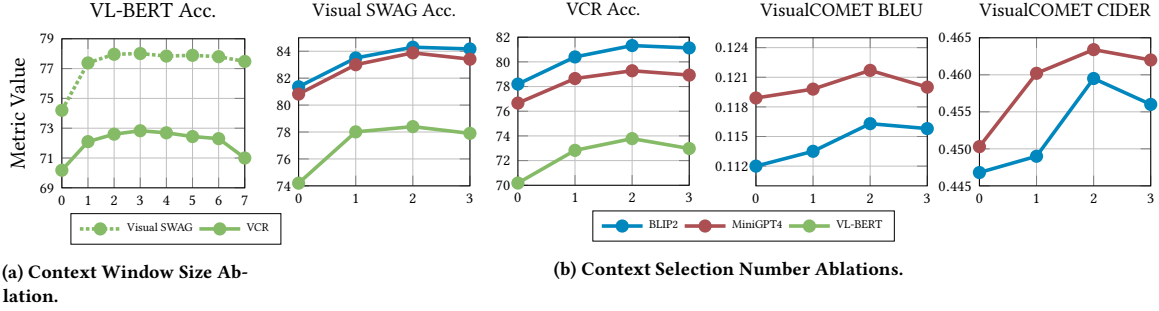
\begin{figure*}[ht]
  \captionsetup{skip=5pt}
  \scalebox{0.9}{
  \begin{subfigure}[t]{0.2\linewidth}%
    \vspace{-\topskip}
    \input{graphs/window_size_selection_num/window_size}
    \vspace{-10pt}
    \caption{Context Window Size Ablation.}
    \label{fig:window_size}
  \end{subfigure}%
  \begin{subfigure}[t]{0.80\linewidth}%
    \label{fig:selection_number}
    \hspace{8pt}
  \begin{minipage}[t]{0.24\linewidth}%
    \vspace{-\topskip}\input{graphs/window_size_selection_num/swag}
  \end{minipage}
    \vspace{-\topskip}\hspace{-10pt}
  \begin{minipage}[t]{0.24\linewidth}%
    \vspace{-\topskip}\input{graphs/window_size_selection_num/vcr}
  \end{minipage}
  \hspace{-10pt}
  \begin{minipage}[t]{0.24\linewidth}%
    \vspace{-\topskip}\input{graphs/window_size_selection_num/vc_b}
  \end{minipage}
  \begin{minipage}[t]{0.24\linewidth}%
    \vspace{-\topskip}\hspace{-6pt}\input{graphs/window_size_selection_num/vc_c}
  \end{minipage}
    \caption{Context Selection Number Ablations.}
    \label{fig:selection_number}
  \end{subfigure}
  }
  \caption{(a) Ablation study for context window size. The performance of VL-BERT in accuracy is plotted on y-axis against context window size on x-axis for Visual SWAG (dotted) and VCR (solid). The peak accuracy at window size 3 indicates the optimal size. (b) Ablation study for context selection number. The performance of multiple models (in different colors) is plotted on y-axis against context selection number on x-axis for multiple datasets. We allow the VLM to observe multiple contexts within the window size. The peak in the curve indicates selecting 2 out of 3 contexts results in the best performance. }
\end{figure*}

%% file: graphs/window_size_selection_num/window_size.tex
\begin{tikzpicture}
\begin{axis}[
    width=1.2\textwidth,
    height=1.06\textwidth,
    title={VL-BERT Acc.},
    ylabel={Metric Value},
    xmin=0, xmax=7,
    ymin=69, ymax=79,
    xtick={0,1,...,7},
    ytick={69,71,...,85},
    grid=both,
    legend style={at={(1.2,-0.17)}, font=\tiny,legend columns=-1},
    ticklabel style = {font=\scriptsize}
]

    \definecolor{VL-BERT}{rgb}{0.52, 0.73, 0.4}
    \definecolor{swag}{rgb}{0.0, 0.53, 0.74}
    \definecolor{vcr}{rgb}{0.67, 0.31, 0.32}
    \addplot[color=VL-BERT,mark=*,mark options={solid}, mark size=2pt,line width=1.5pt, densely dotted] coordinates {
        (0,74.2)
        (1,77.38)
        (2,77.96)
        (3,78.01)
        (4,77.84)
        (5,77.89)
        (6,77.80)
        (7,77.48)
    };
    \addlegendentry{Visual SWAG}
    
    \addplot[color=VL-BERT,mark=*, mark size=2pt,line width=1.5pt] coordinates {
        (0,70.18)
        (1,72.1)
        (2,72.6)
        (3,72.83)
        (4,72.69)
        (5,72.44)
        (6,72.30)
        (7,71)
    };
    \addlegendentry{VCR}
\end{axis}
\end{tikzpicture}
\label{fig:window_size}

%% file: graphs/window_size_selection_num/swag.tex
\begin{tikzpicture}
\begin{axis}[
    width=1.2\textwidth,
    height=1.1\textwidth,
    title={\small Visual SWAG Acc.},
    xmin=0, xmax=3,
    ymin=74, ymax=85,
    xtick={0,1,...,6},
    ytick={74,76,...,85},
    grid=both,legend style={at={(1,-0.3)}, font=\tiny},
    ticklabel style = {font=\scriptsize}
]
    \definecolor{BLIP2}{rgb}{0.0, 0.53, 0.74}
    \definecolor{MiniGPT4}{rgb}{0.67, 0.31, 0.32}
    \definecolor{VL-BERT}{rgb}{0.52, 0.73, 0.4}
    \addplot[color=BLIP2,mark=*, mark size=2pt,line width=1.5pt] coordinates {
        (0,81.36)
        (1,83.5)
        (2,84.30)
        (3,84.17)
    };
    
    \addplot[color=MiniGPT4,mark=*, mark size=2pt,line width=1.5pt] coordinates {
        (0,80.82)
        (1,83.0)
        (2,83.88)
        (3,83.41)
    };
    \addplot[color=VL-BERT,mark=*, mark size=2pt,line width=1.5pt] coordinates {
        (0,74.2)
        (1,78.01)
        (2,78.4)
        (3,77.9)
    };
    \
\end{axis}
\end{tikzpicture}
\label{fig:selection_number}

%% file: graphs/window_size_selection_num/vcr.tex
\begin{tikzpicture}
\begin{axis}[
    width=1.2\textwidth,
    height=1.1\textwidth,
    title={\small VCR Acc.},
    xmin=0, xmax=3,
    ymin=70, ymax=82,
    xtick={0,1,...,6},
    ytick={70,72,...,82},
    grid=both,legend style={at={(2.2,-0.17)}, font=\tiny,legend columns=-1},
    ticklabel style = {font=\scriptsize}
]
    \definecolor{BLIP2}{rgb}{0.0, 0.53, 0.74}
    \definecolor{MiniGPT4}{rgb}{0.67, 0.31, 0.32}
    \definecolor{VL-BERT}{rgb}{0.52, 0.73, 0.4}
    \addplot[color=BLIP2,mark=*, mark size=2pt,line width=1.5pt] coordinates {
        (0,78.20)
        (1,80.40)
        (2,81.32)
        (3,81.13)
    };
  \addlegendentry{BLIP2};
    
    \addplot[color=MiniGPT4,mark=*, mark size=2pt,line width=1.5pt] coordinates {
        (0,76.66)
        (1,78.65)
        (2,79.28)
        (3,78.93)
    };
  \addlegendentry{MiniGPT4};
    \addplot[color=VL-BERT,mark=*, mark size=2pt,line width=1.5pt] coordinates {
        (0,70.18)
        (1,72.83)
        (2,73.78)
        (3,72.98)
    };
  \addlegendentry{VL-BERT};
\end{axis}
\end{tikzpicture}
\label{fig:selection_number}

%% file: graphs/window_size_selection_num/vc_b.tex
\begin{tikzpicture}
\begin{axis}[
    width=1.2\textwidth,
    height=1.1\textwidth,
    title={\small VisualCOMET BLEU},
    xmin=0, xmax=3,
    ymin=0.11, ymax=0.125,
    xtick={0,1,...,6},
    ytick={0.10,0.103,...,0.31},grid=both,
    ticklabel style = {font=\scriptsize},
    y tick label style={
        /pgf/number format/.cd,
        fixed,
        fixed zerofill,
        precision=3,
        /tikz/.cd
    }
]
    \definecolor{BLIP2}{rgb}{0.0, 0.53, 0.74}
    \definecolor{MiniGPT4}{rgb}{0.67, 0.31, 0.32}
    \addplot[color=BLIP2,mark=*, mark size=2pt,line width=1.5pt] coordinates {
        (0,0.1120)
        (1,0.1135)
        (2,0.1163)
        (3,0.1158)
    };
    
    \addplot[color=MiniGPT4,mark=*, mark size=2pt,line width=1.5pt] coordinates {
        (0,0.1189)
        (1,0.1198)
        (2,0.1217)
        (3,0.1200)
    };
\end{axis}
\end{tikzpicture}
\label{fig:selection_number}

%% file: graphs/window_size_selection_num/vc_c.tex
\begin{tikzpicture}
\begin{axis}[
    width=1.2\textwidth,
    height=1.1\textwidth,
    title={\small VisualCOMET CIDER},
    xmin=0, xmax=3,
    ymin=0.445, ymax=0.465,
    xtick={0,1,...,6},
    ytick={0.445,0.450,...,0.49},
    grid=both, 
    ticklabel style = {font=\scriptsize},
    y tick label style={
        /pgf/number format/.cd,
        fixed,
        fixed zerofill,
        precision=3,
        /tikz/.cd
    }
]
    \definecolor{BLIP2}{rgb}{0.0, 0.53, 0.74}
    \definecolor{MiniGPT4}{rgb}{0.67, 0.31, 0.32}
    \addplot[color=BLIP2,mark=*, mark size=2pt,line width=1.5pt] coordinates {
        (0,0.4468)
        (1,0.4490)
        (2,0.4595)
        (3,0.4560)
    };
    
    \addplot[color=MiniGPT4,mark=*, mark size=2pt,line width=1.5pt] coordinates {
        (0,0.4503)
        (1,0.4602)
        (2,0.4634)
        (3,0.4620)
    };
  
\end{axis}
\end{tikzpicture}
\label{fig:selection_number}

%% file: tables/context_modality.tex
\begin{table}[h!]
    \small
    \captionsetup{skip=3pt}
    \caption{(a) Ablation of context and selection modality on Visual SWAG. Best performance is achieved when both context and selection modality is textual. (b) Ablation of context selection methods. The top half selects context using language models while The bottom half selects using context index.}
    \fontsize{8}{9}\selectfont
    \setlength{\tabcolsep}{3pt}
    \begin{subtable}{.5\linewidth}
    \captionsetup{skip=5pt}
    \caption{Context and Selection Modality}
        \fontsize{8}{9}\selectfont
    \setlength{\tabcolsep}{3pt}
    \begin{tabular}{l|l|c}
        \toprule
        Context & Selection & Acc. \\
        \midrule
        No context  & N/A        & 74.20\\
        \midrule
        Text        & Text       & \textbf{78.01}\\
        Text        & Image      & 77.87 \\
        Text        & Text+Image & 77.78 \\
        \midrule
        Image       & Text       & 75.28 \\
        Image       & Image       & 74.97 \\
        \midrule
        Image+Text  & Text       & 77.57 \\
        Image+Text       & Image       & 76.43 \\
        \bottomrule
    \end{tabular}
    \label{table:multimodal_context}
    \end{subtable}%
    \begin{subtable}{.5\linewidth}
    \captionsetup{skip=5pt}
    \caption{Selection Strategy}
        \fontsize{8}{9}\selectfont
    \setlength{\tabcolsep}{3pt}
    \begin{tabular}{l|l|l}
        \hline
        Method & V. SWAG & VCR \\
        \midrule
        Embedding Sim. &76.91 &72.44\\
        Prob Selection & \textbf{78.4} &\textbf{73.78}\\
        \midrule
        Random & 76.79 &71.43\\
        Index -1 & 77.86 & 71.76\\
        Index -2 & 77.25 & 71.01\\
        Index -3 & 76.15 & 70.23\\
        \hline
    \end{tabular}
    \label{table:selection_method}
    \end{subtable} 
\end{table}

%% file: tables/context_model_result.tex
\begin{table}[h!]\small
\captionsetup{skip=5pt}
    \caption{Experiment results of VLMs on Visual SWAG, VCR, and VisualCOMET with/without context. Models with Prob Selection (Prob.) show significant improvement over the baselines. VL-BERT cannot be trained for generative tasks, so results on VisualCOMET are not shown.}
    \centering
    \fontsize{8}{9}\selectfont
    \setlength{\tabcolsep}{3pt}
    \begin{tabular}{l|c|c|ccc}
    \toprule
        \multirow{2}{*}{Model}& V.SWAG&VCR&\multicolumn{3}{c}{VisualCOMET} \\
        &Acc.&Acc.&BLEU4&CIDER&METEOR\\
        \midrule
        VL-BERT & 74.20 & 70.18 & - & - & - \\
        VL-BERT+Prob.& \textbf{78.40} & \textbf{73.78} & - &- &-\\
        \midrule
        BLIP&62.65&69.03&0.1098 &0.4468 &0.1656 \\
        BLIP+Prob.&\textbf{63.22}&\textbf{70.74}&\textbf{0.1147}&\textbf{0.4595}&\textbf{0.1674} \\
        \midrule
        BLIP2 &81.30&78.20&0.1120 &0.4492&0.1648 \\
        BLIP2+Prob.&\textbf{84.36}&\textbf{81.32}&\textbf{0.1163}&\textbf{0.4612}&\textbf{0.1672} \\
        \midrule
        OFA& 54.07 & 69.35&0.1329&0.4446&0.1527    \\
        OFA+Prob.& \textbf{59.44}& \textbf{73.20}& \textbf{0.1354}& \textbf{0.4642}& \textbf{0.1558} \\
        \midrule
        MiniGPT-4 & 80.82 &76.66 &0.1189 &0.4503 &0.1653\\
        MiniGPT-4+Prob.& \textbf{83.88}&\textbf{79.28}&\textbf{0.1217}&\textbf{0.4634}&\textbf{0.1679}\\
    \bottomrule

    \end{tabular}
    \label{table:context_model_result}
\end{table}

%% file: tables/case.tex
\begin{table}[!htbp]\small
\captionsetup{skip=5pt}
    \caption{Performance Analysis of CARA on CASE. While performing better than baselines, CARA also generalizes to new benchmarks without ever been trained on them. 
    }
    \centering
    \fontsize{8}{9}\selectfont
    \setlength{\tabcolsep}{3pt}

\begin{tabular}{l|c|cc}
\toprule
Method                & \begin{tabular}[c]{@{}c@{}}Pseudolabelled\\ Data Source\end{tabular} & \multicolumn{1}{l}{CASE-VCR} & \multicolumn{1}{l}{CASE-V.SWAG} \\ \hline
\multirow{2}{*}{CARA} & VCR                                                                   & \textbf{75.69}               & 64.55                           \\
                      & V.SWAG                                                                & 54.09                        & \textbf{73.05}                  \\
Selector-MaxProb      & -                                                                     & 51.03                        & 50.1                            \\
Selector-MLP          & -                                                                     & 54.82                        & 53.84                           \\ 
\bottomrule
\end{tabular}
    \label{table:case}
\end{table}

%% file: tables/abstain.tex
\begin{table}[h!]\small
\captionsetup{skip=5pt}
    \caption{VLM performance enhancement by using CARA. CARA consistently improves performance across VLMs and benchmarks. * indicates the performance obtained via applying CARA trained on VCR.}
    \centering
    \fontsize{8}{9}\selectfont
    \setlength{\tabcolsep}{3pt}
    \begin{tabular}{l|l|c|ccc}
    \toprule
        \multirow{2}{*}{Model}&V.SWAG&VCR&\multicolumn{3}{c}{VisualCOMET}\\
        &Acc.&Acc.&BLEU4&CIDER&METEOR\\
        \midrule
        VL-BERT&73.72&70.13 & -&-&-\\
        VL-BERT+CARA &\textbf{77.04 (74.76*)}&\textbf{73.40}&-&-&-\\
        \midrule
        BLIP2&81.30&78.20&0.1120&0.4492&0.1648\\
        BLIP2+CARA &\textbf{82.93}&\textbf{79.77}&\textbf{0.1179*}&\textbf{0.4642*}&\textbf{0.1674*}\\
    \bottomrule

    \end{tabular}
    \label{table:abstention1}
\end{table}

%% file: tables/abstain_generalization_v1.tex
\begin{table}[h!]
    \small
    \captionsetup{skip=5pt}
    \caption{CARA Generalization. CARA improves VLM performance across several benchmarks despite never being trained on them, indicating its utility even for future benchmarks. }
    \centering
    \fontsize{8}{9}\selectfont
    \setlength{\tabcolsep}{0.9pt}
\begin{tabular}{l|cc|cc|cc|cc}
\toprule
\multirow{2}{*}{Model}                     & \multicolumn{2}{c|}{VQA v2}                       & \multicolumn{2}{c|}{GQA}                          & \multicolumn{2}{c|}{OKVQA}                        & \multicolumn{2}{c}{A-OKVQA}                       \\
                                           & \multicolumn{1}{l}{} & \multicolumn{1}{l|}{+CARA} & \multicolumn{1}{l}{} & \multicolumn{1}{l|}{+CARA} & \multicolumn{1}{l}{} & \multicolumn{1}{l|}{+CARA} & \multicolumn{1}{l}{} & \multicolumn{1}{l}{+CARA} \\ \midrule
\textit{Zero-shot}&&&&&&& \\
BLIP2& 62.5                 & \textbf{64.9}              & 46.33                & \textbf{47.57}             & 34.68                & \textbf{36.55}             & 43.94                & \textbf{45.00}            \\
PNP & 57.52                & \textbf{60.42}             & 35.68                & \textbf{36.96}             & 26.98                & \textbf{28.54}             & 27.78                & \textbf{28.43}            \\ \midrule
Prophet& -                    & -                          & -                    & -                          & 61.1                 & \textbf{62.28}             & 58.20                & \textbf{58.37}            \\
PromptCap & -                    & -                          & -                    & -                          & 60.44                & \textbf{61.54}             & 60.43                & \textbf{60.59}            \\
MCAN& -                    & -                          & -                    & -                          & 53.05                & \textbf{53.63}             & 51.97                & \textbf{52.09}            \\ 
\bottomrule
\end{tabular}
    \label{table:abstain_gen1}
\end{table}

%% file: tables/abstain_generalization_v2.tex
\begin{table}[h!]
    \small
    \captionsetup{skip=5pt}
    \caption{
    Analysis of CARA abstained samples by humans, with percentages indicating "Abstained" samples where the model refrained from predicting, and "Ambiguous" and "Insufficient" denoting the proportions of abstained samples judged as such. Samples lacking context are considered ambiguous, but not vice versa. Majority of CARA-abstained samples are ambiguous, proving CARA works by removing ambiguous samples, not hard samples. 
    }
    \centering
    \fontsize{8}{9}\selectfont
    \setlength{\tabcolsep}{2pt}

\begin{tabular}{l|c|cccc}
\toprule
                             & \multicolumn{1}{l|}{Abstention} & \multicolumn{1}{l}{VQA v2 } & \multicolumn{1}{l}{GQA} & \multicolumn{1}{l}{OKVQA} & \multicolumn{1}{l}{A-OKVQA} \\ \midrule
Abstained            & \multirow{3}{*}{CARA}           & 10.90                      & 5.78                    & 28.77                     & 5.12                       \\
Ambiguous&                                 & 69.00                      & 70.00                   & 64.00                     & 69.00                      \\
Insufficient Context &                                 & 47.00                      & 42.00                   & 46.00                     & 53.00                      \\ \midrule
Abstained            & \multirow{3}{*}{Selector-MLP}   & 21.07                      & 17.05                   & 34.08                     & 25.08                      \\
Ambiguous  &                                 & 23.00                      & 16.00                   & 25.00                     & 17.00                      \\
Insufficient Context &                                 & 18.00                      & 14.00                   & 20.00                     & 16.00                      \\ 
\bottomrule
\end{tabular}
    \label{table:abstain_gen2}
\end{table}

%% file: tables/risk.tex
\begin{table*}[!h]\small
    \centering
    \fontsize{8}{9}\selectfont
    \setlength{\tabcolsep}{3pt}
    \captionsetup{skip=5pt}
    \definecolor{mediumblue}{rgb}{0.0, 0.0, 0.8}
    \definecolor{utahcrimson}{rgb}{0.83, 0.0, 0.25}
    \caption{Abstention analysis on VQA v2. We compare CARA's system risk $\mathcal{R}$, effective reliability $\Phi_1$, and coverage $\mathcal{C}$ against baselines. The arrows following the metrics indicate the direction of improvement (for example $(\downarrow)$ indicates lower the better). Risks in \textcolor{utahcrimson}{red} are higher than tolerance, and metrics are highlighted in \textcolor{mediumblue}{blue} when CARA outperforms the baseline methods.}
    \begin{tabular}{l|l|ccc|ccc|ccc}
        \toprule
        \multirow{2}{*}{VLM} & \multirow{2}{*}{Method} & \multicolumn{3}{c|}{Risk Tolerance $r=10\%$}& \multicolumn{3}{c|}{Risk Tolerance $r=30\%$}&\multicolumn{3}{c}{Risk Tolerance $r=50\%$} \\
        &&$\mathcal{R}(\downarrow)$&$\Phi_1(\uparrow)$& $\mathcal{C}(\uparrow)$&$\mathcal{R}(\downarrow)$&$\Phi_1(\uparrow)$& $\mathcal{C}(\uparrow)$&$\mathcal{R}(\downarrow)$&$\Phi_1(\uparrow)$& $\mathcal{C}(\uparrow)$\\
        \midrule
        \multirow{3}{*}{BLIP2} & Selector-MaxProb&
        1.4&5.0&5.2&4.9&18.7&21.2&8.5&29.3&36.8\\
        &Selector-MLP& 2.3&7.9&8.7&7.4&25.1&30.2&11.7&36.0&46.2\\
        &CARA&2.6&\textcolor{mediumblue}{10.0}&\textcolor{mediumblue}{10.7}&8.8&\textcolor{mediumblue}{31.1}&\textcolor{mediumblue}{38.8}&13.1&\textcolor{mediumblue}{39.0}&\textcolor{mediumblue}{56.3}\\
        \midrule
        \multirow{3}{*}{PNP} & Selector-MaxProb&\textcolor{utahcrimson}{29.5}&2.3&7.3&\textcolor{utahcrimson}{35.5}&12.1&45.8&41.6&23.2&79.5\\
        &Selector-MLP& \textcolor{utahcrimson}{29.0}&9.8&11.6&\textcolor{utahcrimson}{35.4}&23.5&48.9&41.4&29.8&78.5\\
        &CARA&\textcolor{utahcrimson}{28.9}&\textcolor{mediumblue}{13.6}&\textcolor{mediumblue}{15.4}&\textcolor{utahcrimson}{36.6}&\textcolor{mediumblue}{28.6}&\textcolor{mediumblue}{51.3}&41.8&\textcolor{mediumblue}{33.5}&\textcolor{mediumblue}{80.1}\\
        \bottomrule

    \end{tabular}
    \label{table:risk_coverage}
\end{table*}

%% file: graphs/qualitative_examples.tex
\definecolor{chosen_context}{rgb}{0.4752, 0.6273, 0.84}
\definecolor{incorrect_answer}{rgb}{0.7, 0.13, 0.13}
\definecolor{correct_answer}{rgb}{0.52, 0.73, 0.4}

\begin{figure}[h!]
    \centering
    \captionsetup{skip=5pt}
    \includegraphics[width=\linewidth]{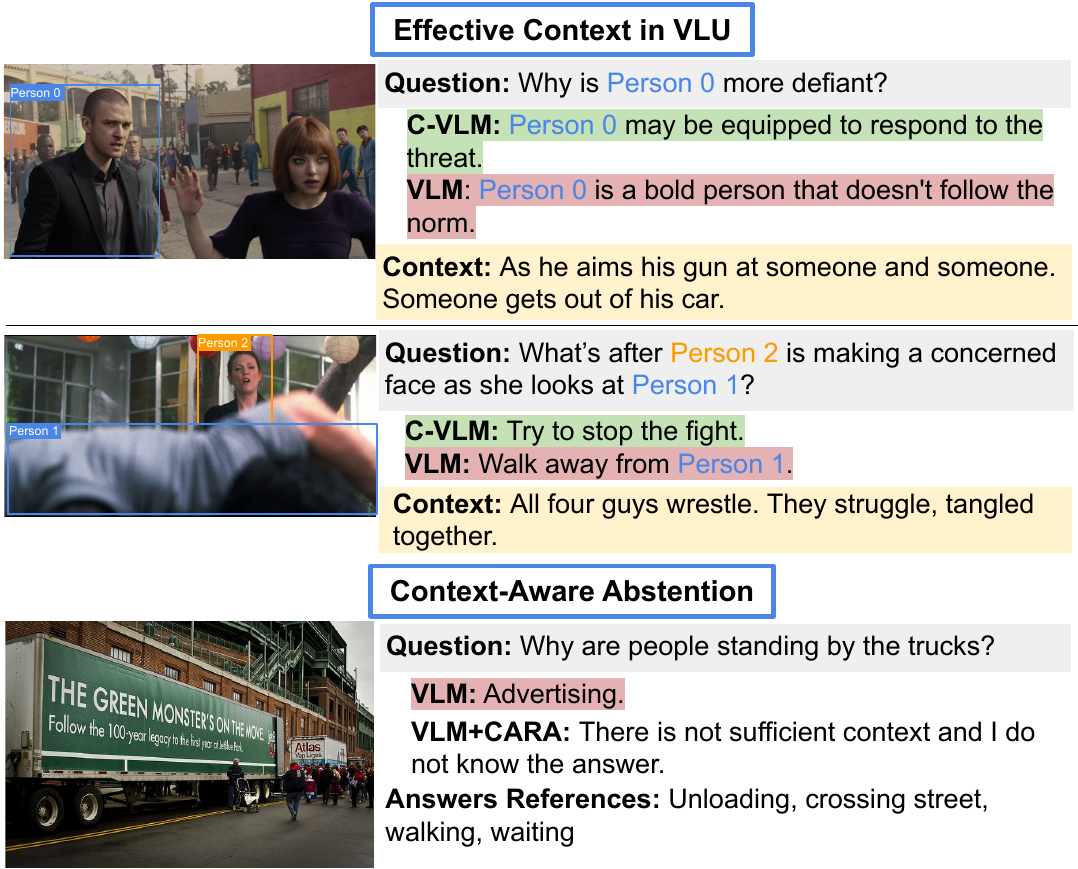}
  \caption{Qualitative Examples. Correct answers are highlighted in \colorbox{correct_answer!50}{green}. Incorrect answers are highlighted in \colorbox{incorrect_answer!35}{red}.}
  \label{fig:qualitative_examples}
  
\end{figure}

%% file: graphs/context_example.tex
\begin{figure}[h!]
    \centering
    \includegraphics[width=\linewidth]{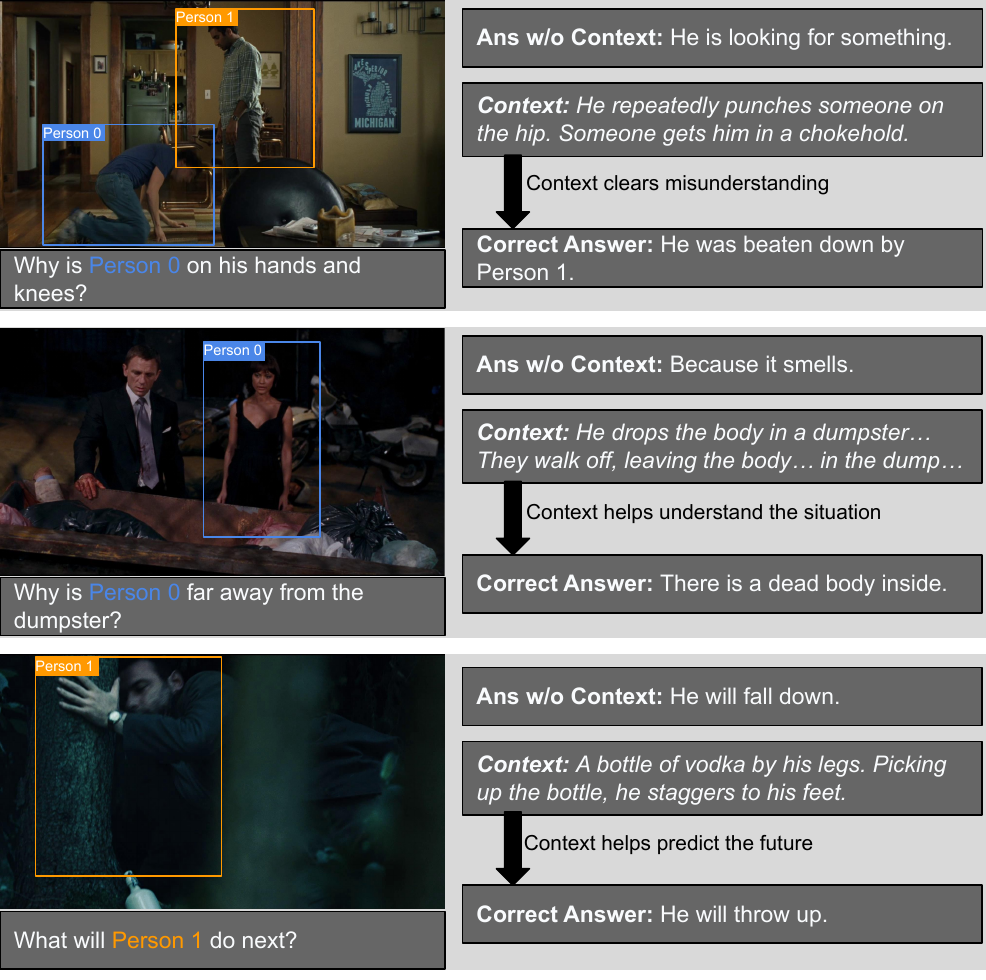}
    \caption {Three scenarios of how context can help understanding in an image-language reasoning task. In the first row, The fighting scene in the context suggests he is down because of the injury, but not what he seems to be doing in the image. In the second row, the context mentions the presence of a corpse invisible in the image, so the woman is more likely to stay away because of fear instead of distaste. In the third row, The appearance of a vodka bottle and his stumbling indicate he is drunk, which makes the correct answer more plausible.}
    \label{fig:context_importance}
\end{figure}

%% file: graphs/cara-verification.tex
\begin{figure}[t]
    \centering
    \includegraphics[width=\linewidth]{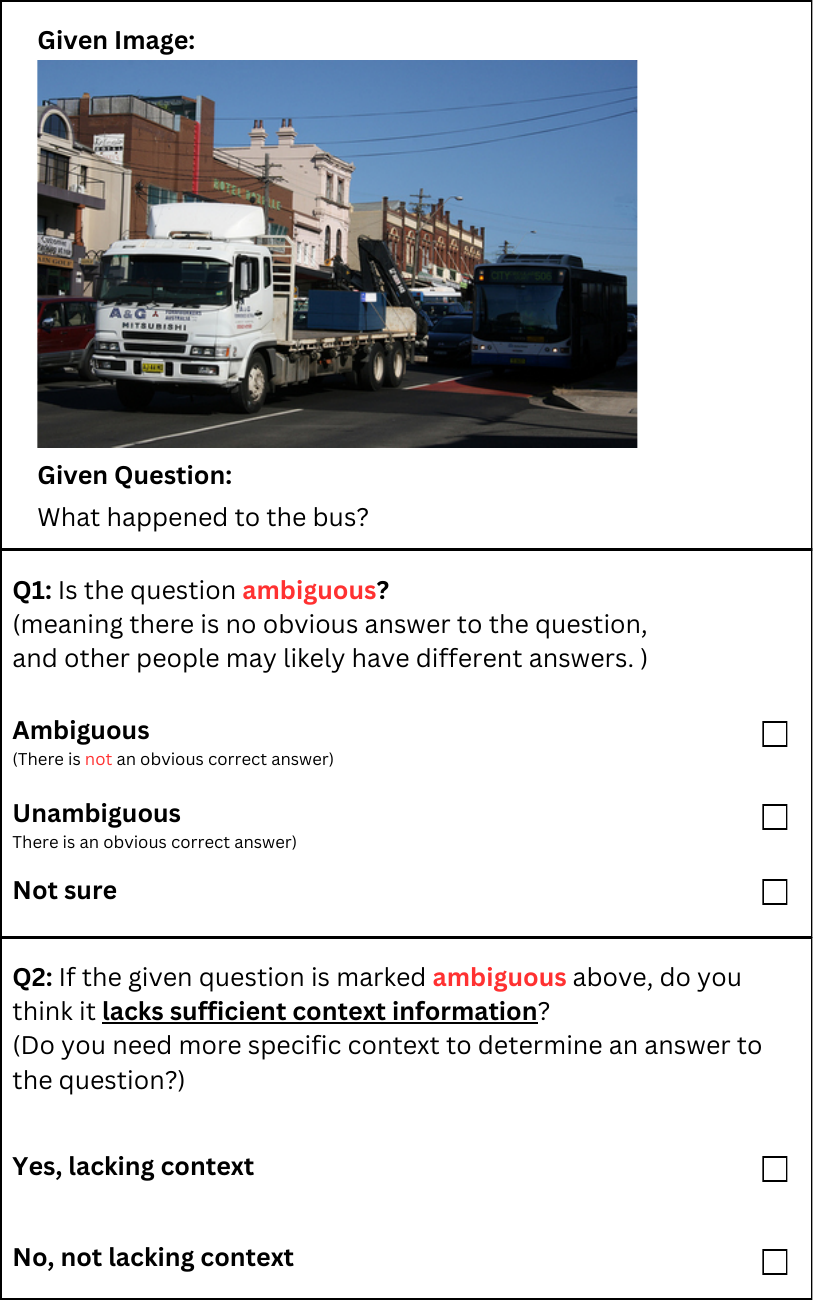}
  \caption{Interface layout for annotators in verifying the correctness of CARA's detection results. We implemented this interface over the Amazon Turker platform to facilitate turkers to effectively understand the assignment and annotate the data. In practice, we also include plenty of annotated examples beforehand as the instruction or reference.}
  \label{fig:cara-verify}
\end{figure}

%% file: graphs/context_retrieval.tex
\begin{figure}[h!]
    \centering
    \captionsetup{skip=5pt}
    \includegraphics[width=\linewidth]{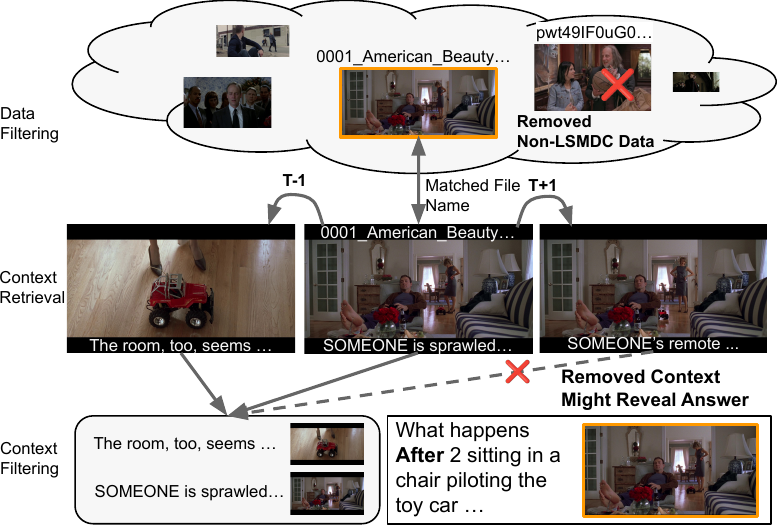}
  \caption{Dataset Construction Process: 1. Remove non-LSMDC data points. 2. Find the source clip for each image by matching the file names and save the corresponding captions as context. 3. Filter out context that can potentially reveal the correct answer.}
  \label{fig:context_retrieval}
\end{figure}

%% file: tables/abstain_generalization_appendix.tex
\begin{table*}[h!]
    \small
    \captionsetup{skip=5pt}
    \caption{
    Analysis of CARA abstained samples by humans, with percentages indicating "Abstained" samples where the model refrained from predicting, and "Ambiguous" and "Insufficient" denoting the proportions of abstained samples judged as such. Samples lacking context are considered ambiguous, but not vice versa. Majority of CARA-abstained samples are ambiguous, proving CARA works by removing ambiguous samples, not hard samples. 
    }
    \centering
    \fontsize{8}{9}\selectfont
    \setlength{\tabcolsep}{2pt}

\begin{tabular}{l|c|ccccccc}
\toprule& Abstention& VCR & VisualCOMET & Visual SWAG&VQA v2&GQA&OKVQA&A-OKVQA\\ \midrule
Abstained& \multirow{3}{*}{CARA}&13.66&18.14&18.73& 10.90 & 5.78& 28.77& 5.12\\
Ambiguous&&88.00&98.00&78.00&69.00&70.00& 64.00& 69.00\\
Insufficient Context&&82.00&98.00&74.00& 47.00&42.00& 46.00& 53.00\\ \midrule

Abstained& \multirow{3}{*}{Selector MLP}&24.83&25.90&24.08&21.07&17.05&34.08& 25.08\\
Ambiguous&&58.00&72.00&65.00&23.00&16.00&25.00& 17.00\\
Insufficient Context&&32.00&70.00&58.00&18.00&14.00&20.00&16.00\\ \midrule
\end{tabular}
    \label{table:abstain_gen_appendix}
\end{table*}

%% file: graphs/appendix_qualitative_example.tex
\definecolor{cap_chosen}{rgb}{0.4752, 0.6273, 0.84}
\definecolor{no_cap_choice}{rgb}{0.7, 0.13, 0.13}
\definecolor{cap_choice}{rgb}{0.52, 0.73, 0.4}
\begin{figure*}[h!]
    \includegraphics[width=\linewidth]{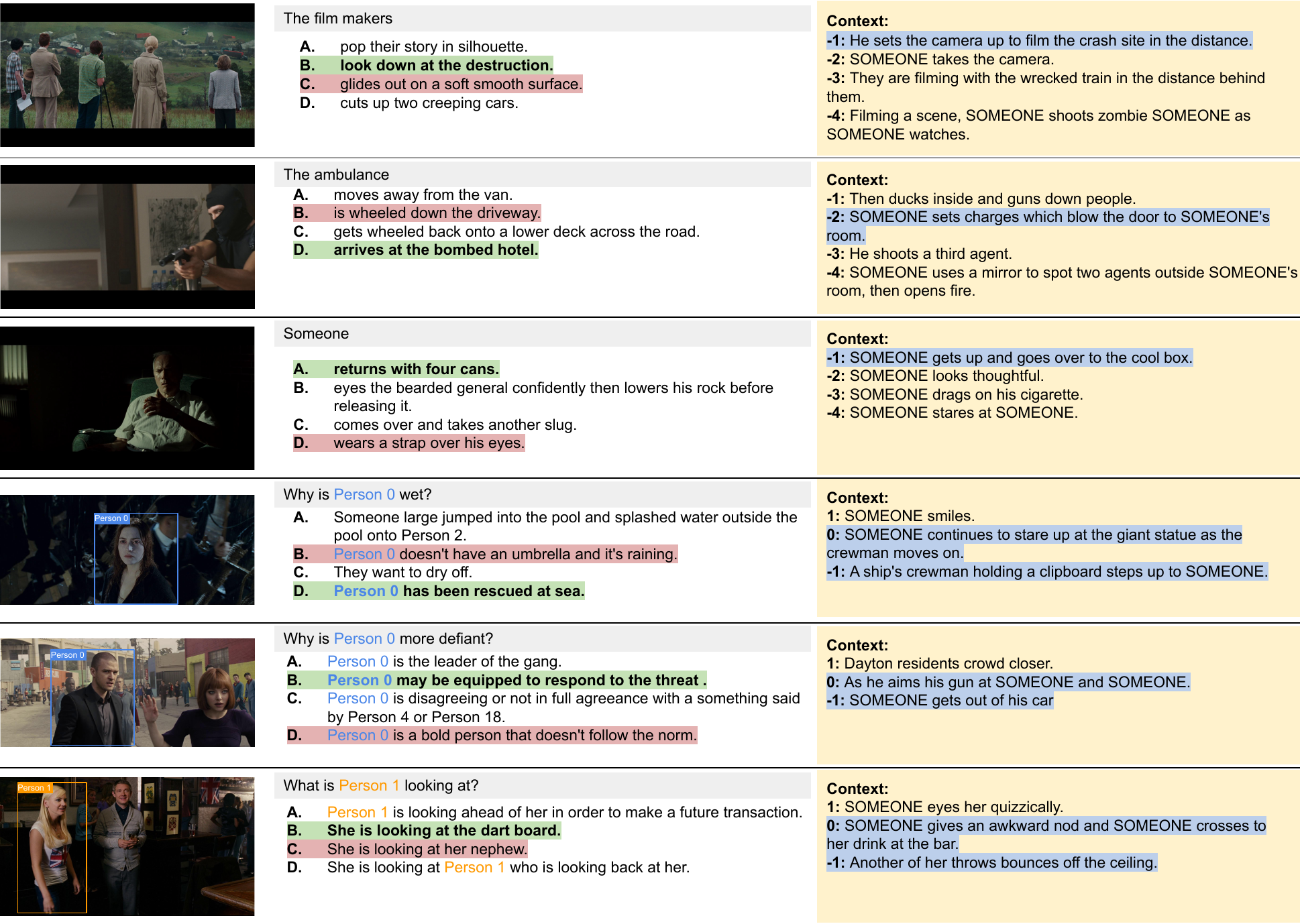}
  \caption{Qualitative examples of Visual SWAG (example 1-3) and VCR (4-6) with/without context. Predictions made by context model are highlighted in \colorbox{cap_choice!50}{Green}. Predictions made by no context models are highlighted in \colorbox{no_cap_choice!35}{Red}. The selected context is highlighted in \colorbox{cap_chosen!50}{Blue}. Correct choices are in \textbf{Bold} font for Visual SWAG and VCR examples.}
  \label{fig:q_examples_swag_vcr}
\end{figure*}

\begin{figure*}[h!]

    \includegraphics[width=\linewidth]{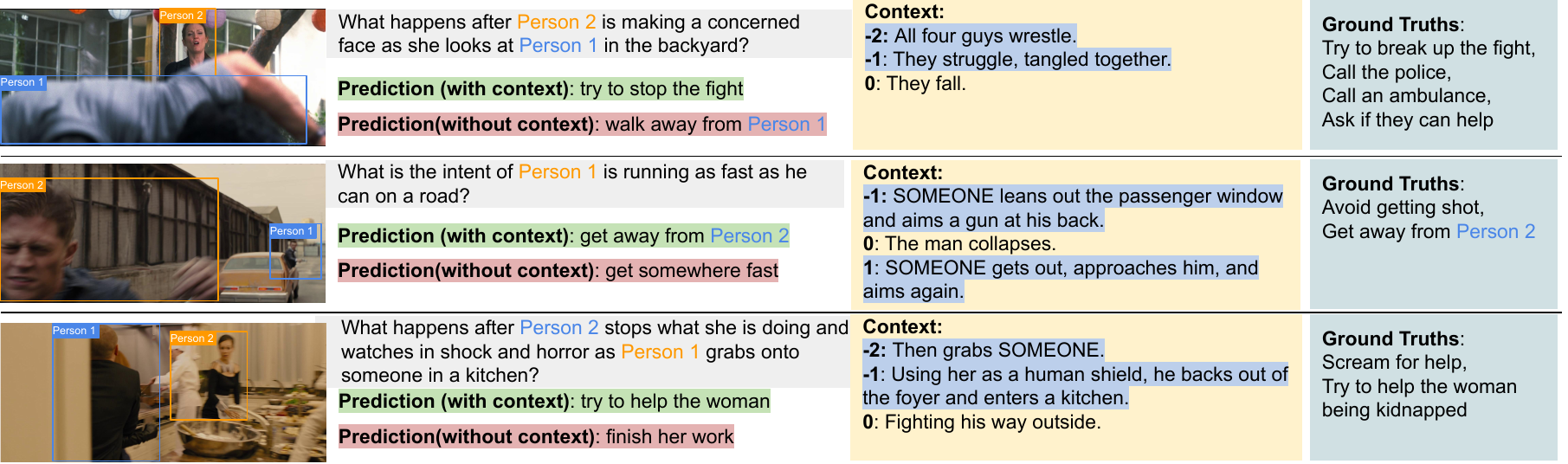}
  \caption{Qualitative examples of VisualCOMET with/without context. Predictions made by context model are highlighted in \colorbox{cap_choice!50}{Green}. Predictions made by no context models are highlighted in \colorbox{no_cap_choice!35}{Red}. The selected context is highlighted in \colorbox{cap_chosen!50}{Blue}.}
  \label{fig:q_examples_visualcomet}
\end{figure*}
\begin{figure*}[h!]

    \includegraphics[width=\linewidth]{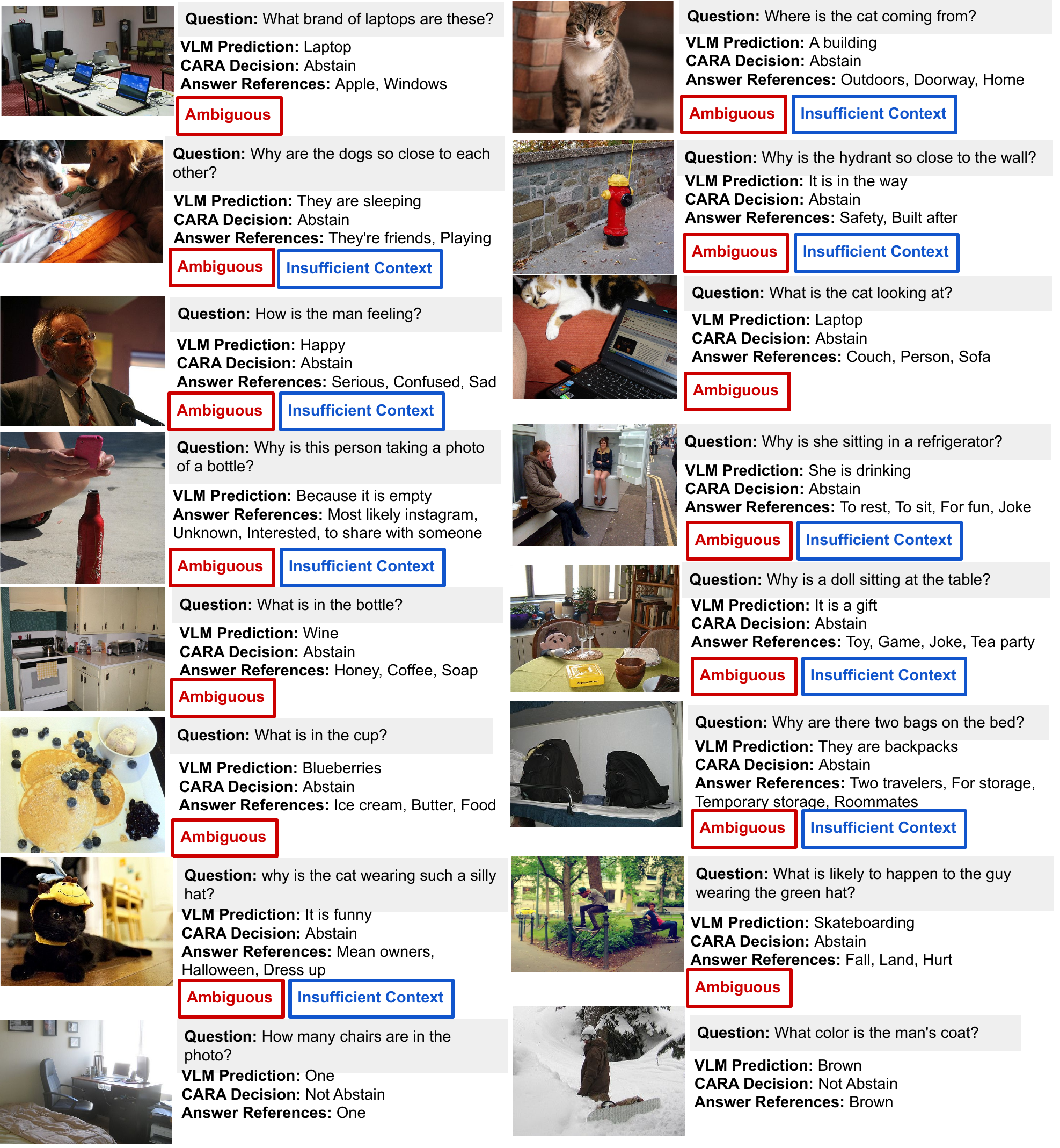}
  \caption{Additional qualitative examples answered by BLIP2. The labels ``Ambiguous" and ``Insufficient Context" under samples abstained by CARA are determined by human annotators. }
  \label{fig:abstained}
\end{figure*}